\documentclass[letterpaper, 10 pt, conference]{ieeeconf}

\IEEEoverridecommandlockouts                       
\usepackage{siunitx}
\usepackage{graphicx}
\usepackage{multirow}
\usepackage{adjustbox}
\usepackage{amsmath}
\usepackage[hyphens]{url}
\usepackage{hyperref}
\usepackage{algorithm}
\usepackage{algpseudocode}
\usepackage[table,xcdraw]{xcolor}

\usepackage{subcaption}

\usepackage{siunitx}
\usepackage{wrapfig}
\usepackage{array}
\usepackage{booktabs}
\usepackage{svg}
\title{\LARGE \bf
Accelerating Real-World Overtaking in F1TENTH Racing Employing Reinforcement Learning Methods}

\author{Emily Steiner$^{1*}$, Daniel van der Spuy$^{1}$, Futian Zhou$^{1}$, Afereti Pama$^{1}$, Minas Liarokapis$^{2}$, Henry Williams$^{1}$\\
\thanks{$^{1}$The authors are with the Centre for Automation and Robotic Engineering Science, The University of Auckland, New Zealand.}%
\thanks{$^{2}$The authors are with New Dexterity research group, The University of Auckland, New Zealand.}
\thanks{Corresponding author: emily.steiner@auckland.ac.nz$^{*}$}
}

\begin{document}

\maketitle
\thispagestyle{empty}
\pagestyle{empty}

\begin{abstract}
   While autonomous racing performance in Time-Trial scenarios has seen significant progress and development, autonomous wheel-to-wheel racing and overtaking are still severely limited. These limitations are particularly apparent in real-life driving scenarios where state-of-the-art algorithms struggle to safely or reliably complete overtaking manoeuvres. This is important, as reliable navigation around other vehicles is vital for safe autonomous wheel-to-wheel racing. The F1Tenth Competition provides a useful opportunity for developing wheel-to-wheel racing algorithms on a standardised physical platform. The competition format makes it possible to evaluate overtaking and wheel-to-wheel racing algorithms against the state-of-the-art. This research presents a novel racing and overtaking agent capable of learning to reliably navigate a track and overtake opponents in both simulation and reality. The agent was deployed on an F1Tenth vehicle and competed against opponents running varying competitive algorithms in the real world. The results demonstrate that the agent's training against opponents enables deliberate overtaking behaviours with an overtaking rate of 87\% compared 56\% for an agent trained just to race.
\end{abstract}
\section{INTRODUCTION}
    Autonomous racing has seen significant developments and improvements over the past five years, with autonomous time-trial racing showcasing times rivalling human drivers \cite{Wurman2022}. However, one area of autonomous racing that is underdeveloped is wheel-to-wheel racing. This can be attributed to various factors, including complexity and the increased risk of damage to expensive equipment. 

    Autonomous racing competitions, such as F1Tenth, provide researchers with opportunities to push the limits of autonomous capabilities \cite{Betz2019}. Wheel-to-wheel racing is an important area of autonomous racing development because it tests autonomous vehicles on how they interact with other vehicles in their vicinity. This has implications not just for autonomous racing but also for autonomous commercial vehicles, which are expected to manoeuvre safely around other vehicles on the road.
    
\begin{figure}[!htp]
        \centering
        \includegraphics[width = 1\linewidth]{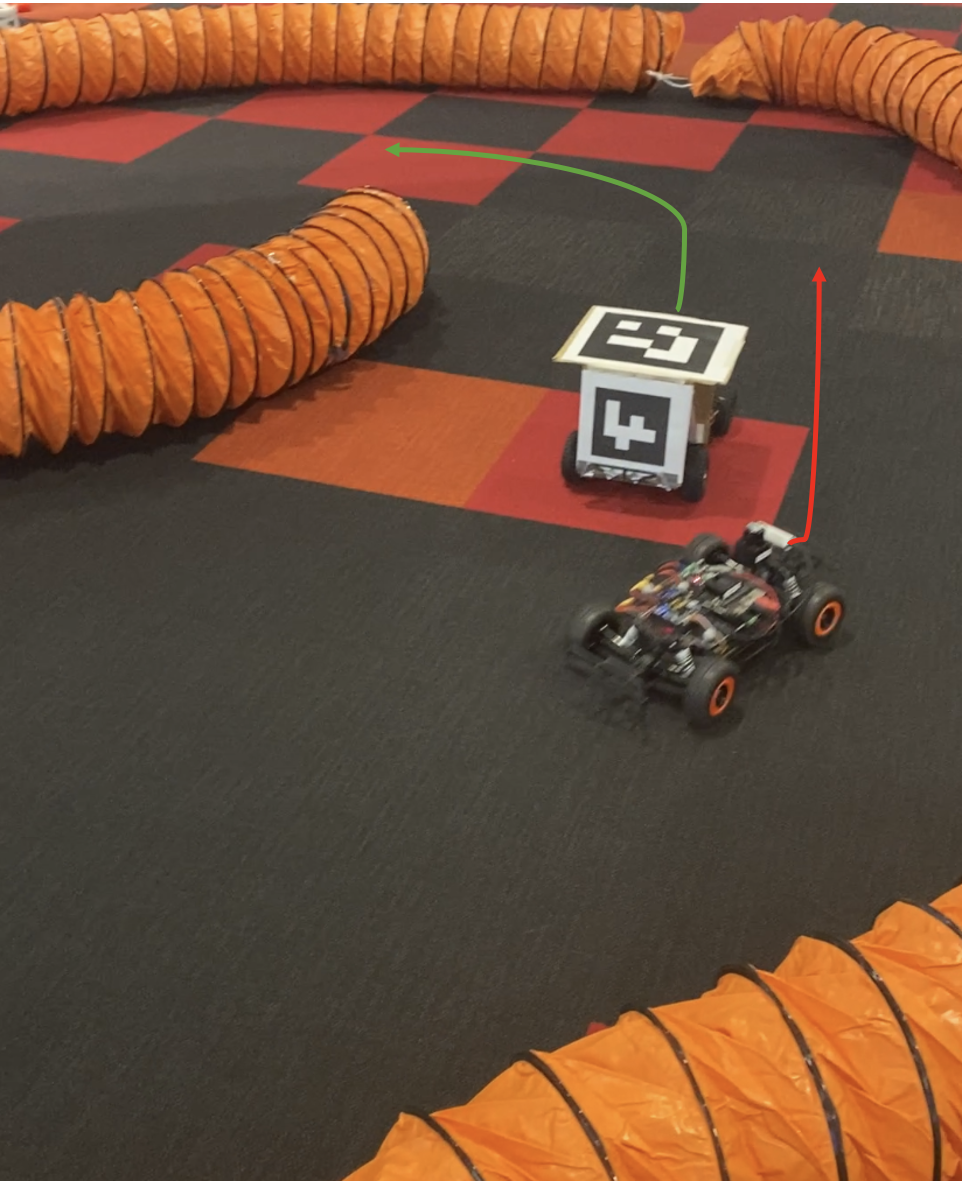}
        \caption{F1Tenth vehicle beginning an overtake manoeuvre on an opponent, moving wide to go around the competition. Full video of overtaking manoeuvres is available at \url{https://youtu.be/6vRRWeZTG-k}}
        \label{fig:f1tenth}
    \end{figure}

    The F1Tenth competition provides a unique opportunity for autonomous racing developers to hold autonomous wheel-to-wheel competitions on racing platforms approximately one-tenth of the size of a standard racing car \cite{OKelly2020}, as shown in Figure \ref{fig:f1tenth}. The competition uses an affordable, standardised vehicle platform, which allows teams to focus primarily on software development for autonomous racing. The affordability of the platforms and components, in addition to the lower racing speeds compared to full-size racecars, allows developers to push boundaries that could not be risked with full-size vehicles. This makes the F1Tenth competition an ideal framework for developing overtaking and adversarial racing techniques for autonomous vehicles.

    The challenge of wheel-to-wheel autonomous racing adds an additional layer of complexity to single-car time trial racing. In addition to navigating the static track at speed, the vehicle must also avoid other dynamic vehicles while modifying its own racing strategy to overtake other vehicles and avoid being overtaken. This introduces significantly more uncertainty to the problem, which makes employing traditional control methods difficult. Instead, control techniques must be highly adaptable to respond to varying competitor racing strategies across various race tracks.

    This work presents a novel autonomous F1Tenth racing strategy with overtaking behaviours learned through reinforcement learning. This strategy is capable of overtaking autonomous racing adversaries on a competition race track. The learned agent shows deliberative overtaking behaviours both in simulation and real life. This is achieved by training the vehicle using reinforcement learning techniques to navigate a previously unknown track, overtake other vehicles on the track, and maintain the lead against such vehicles.

\section{RELATED WORK}

    Autonomous overtaking has been identified as one of the key areas still to be achieved in the autonomous driving literature \cite{Betz2022}. There has been some success in overtaking for passenger vehicles in highly constrained scenarios \cite{Murgovski2015, Perez2011}. However, autonomous overtaking in racing has largely been limited to simulation. 

    Autonomous overtaking methods which have performed well in simulation include Gaussian methods \cite{Bruedigam2021, Schwarting2021}, obstacle avoidance methods \cite{Kalaria2021, Baumann2024}, Neural Networks \cite{Zarrar2024, Zhang2023}, MPC and NMPC \cite{Buyval2017}, and learning methods, such as reinforcement learning \cite{Song2021, Loiacono2010, Huang2015}. Generally, methods that required precise knowledge of the environment struggled with the dynamic nature of overtaking \cite{Dixit2018}, giving more adaptive methods an edge.

    Two studies \cite{SureshBabu2022, Baumann2024} used state machines to implement overtaking methods on real F1Tenth cars. Baumann et al. \cite{Baumann2024} has had success with this method, winning several F1Tenth competitions. However, the exact data on their overtaking success rate is unknown. Meanwhile, Babu et al. \cite{SureshBabu2022} experienced limited success with the method, achieving a success rate of only 39.5\%. Their method gave the ego car information on its opponent's main control frame, and they tested at various relative speeds, including giving both agents the same maximum speed (excluding the boost overtaking speed given to the ego car, which was also varied in the experiments). The overtaking success for each relative speed setting is not provided. One limitation that could have contributed to their low success rate is the lack of distinction between overtaking attempts that led to crashes and overtaking attempts that were then aborted. The number of crashes is not provided.

     The third place at the 12th F1Tenth Autonomous Grand Prix used a TinyLiDARNet method, which uses CNNs on LiDAR information \cite{Zarrar2024}. The end-to-end method provides a LiDAR scan of 1081 points and produces a steering angle and speed. Successful overtaking appears to have been achieved as a side-effect of the method's obstacle-avoidance capabilities. However, similar to \cite{Baumann2024}, they are yet to publish exact data on their overtaking success rate.

    One study that implemented their overtaking on a real full-scale car used game theoretic planning for adversarial racing \cite{Wang2019a}. This method successfully overtook a simulated opponent up to seven times; however, a success rate was not provided. They always used the same track and gave the ego car extra knowledge of the competitor car's odometry - information that would not be given in a racing environment. 

\section{OVERTAKING WITH REINFORCEMENT LEARNING}

    An end-to-end reinforcement learning agent was trained using the autonomous F1Tenth simulator \cite{Steiner2024}, which uses a ROS 2 Humble and Gazebo framework. The simulated learning is designed to replicate the real car as much as possible. The robot can access ROS nodes publishing LiDAR data, linear and angular velocity, and reward information. A novel overtaking environment was developed in the autonomous F1Tenth simulator to train an agent to learn to overtake while racing.

    \subsection{Overtaking Training Environment}
    The training environment pits the training vehicle against four competitors operating with the conventional but competitive racing method of Follow the Gap \cite{Sezer2012, Klapalek2021}. Follow the Gap is an obstacle avoidance method which follows the widest gap in the 'visible' field. It was used to win the F1Tenth competition in 2018 \cite{Klapalek2021}. 
    
\begin{figure*}[!ht]
        \centering
        \begin{subfigure}{0.16\textwidth}
            \centering
            \includegraphics[width=0.9\textwidth]{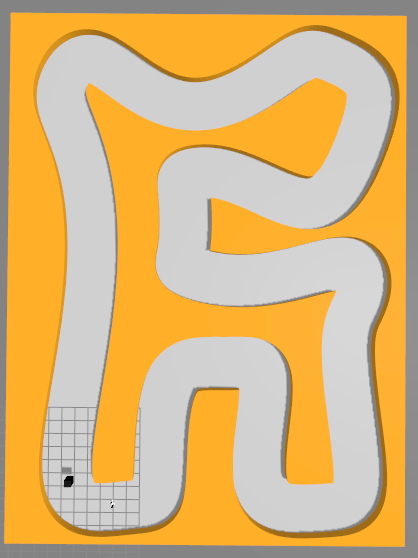}
            \caption{Track 1}
        \end{subfigure}
        \begin{subfigure}{0.16\textwidth}
            \centering
            \includegraphics[width=1\textwidth]{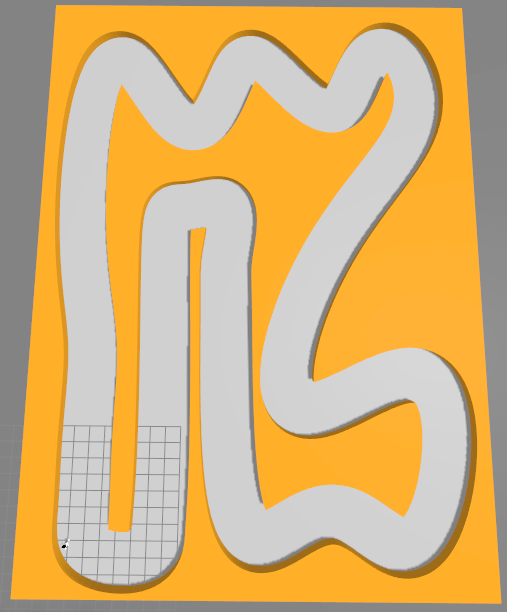}
            \caption{Track 2}
        \end{subfigure}
        \begin{subfigure}{0.16\textwidth}
            \centering
            \includegraphics[width=0.91\textwidth]{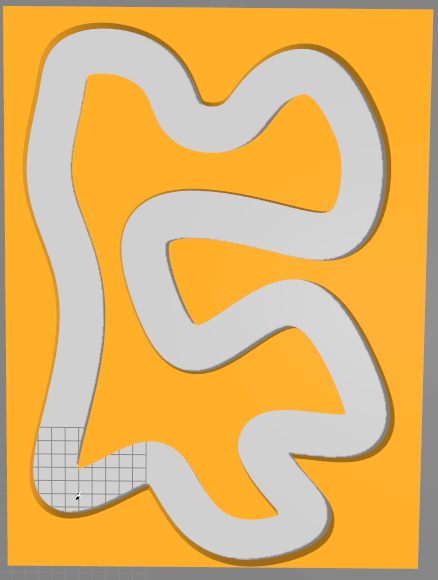}
            \caption{Track 3}
        \end{subfigure}
        \begin{subfigure}{0.16\textwidth}
            \centering
            \includegraphics[width=1\textwidth]{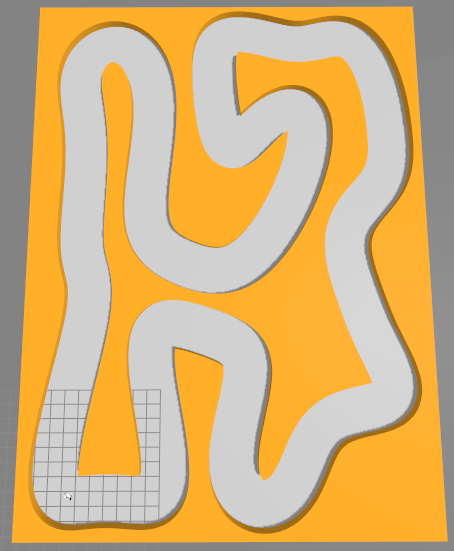}
            \caption{Track 4}
        \end{subfigure}
        \begin{subfigure}{0.16\textwidth}
            \centering
            \includegraphics[width=0.96\textwidth]{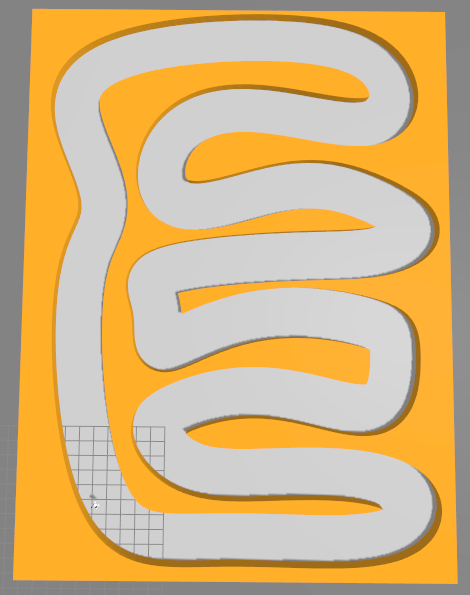}
            \caption{Track 5}
        \end{subfigure}
        \begin{subfigure}{0.16\textwidth}
            \centering
            \includegraphics[width=0.96\textwidth]{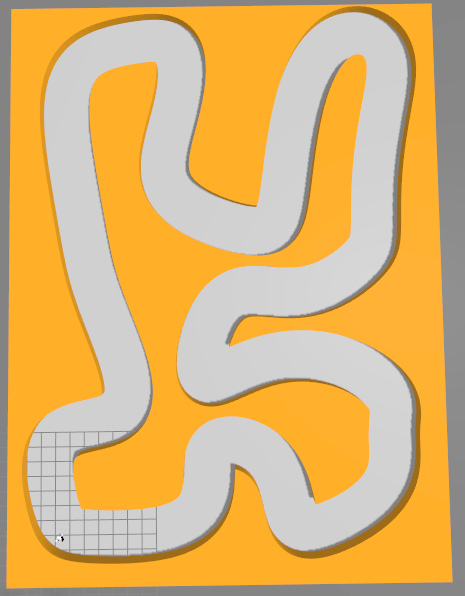}
            \caption{Track 6}
        \end{subfigure}
    
        \caption{Training tracks in simulation. The models were designed in CAD using a spline function, which was then extruded to widths ranging from \SI{1.5}{\meter} to \SI{3.5}{\meter}. The figure shows these models as they are rendered in the Gazebo environment.}
        \label{fig:track}
    \end{figure*}

    Training operates across 24 tracks to provide the agent with a wide range of racing conditions. These tracks were generated by designing six competition legal tracks with various straights and turns (as shown in Figure \ref{fig:track}). These tracks are represented by a spline, which was then extruded to four widths ranging from \SIrange{1.5}{3.5}{\meter} to provide various track widths to learn on. 
    
    Training episodes begin with the ego car spawning at a random way-point along the spline of a random track. The four competitor cars then spawn ahead of the ego car at between 2 and 30 waypoints (the distance between each waypoint is $<$ \SI{1}{\meter}) further along the spline. The cars then start racing, with the ego car running the learning agent with a maximum speed of \SI{2}{\meter\per\sec} and the competitor cars running Follow the Gap at a maximum velocity of \SI{1.5}{\meter\per\sec}. This work intends to focus on learning overtaking behaviours; the difference in maximum velocity enables frequent overtaking opportunities to explore this interaction. A training episode will end if a collision occurs, the ego car fails to progress along the spline for five steps, or the maximum episode steps (3000 steps) are reached. The maximum episode steps setting was implemented to prevent the agent from becoming overly familiar with one track. 
    
    \subsection{State and Action Space}
    The state space was chosen to minimise differences between the simulation and reality and to eliminate reliance on odometry and extra information that would not be available in a competition (such as information on the speed or location of competitor vehicles). The state space consists of the car's current velocity, \(v\), steering angle \(\theta\), and ten LiDAR points \(d_i\) generated from an averaging filter. The LiDAR points' averaging reduces the state space's size while maintaining a practical view of the car's immediate area (within the 270\textdegree{}  detection angle). This strategy of reducing the LiDAR space has shown to be effective when applying reinforcement learning for racing \cite{Evans2023}. The continuous action space controls the vehicle's linear velocity between \SIrange{0}{2}{\meter\per\sec} and steering angle between \SIrange{-0.434}{0.434}{\radian\per\sec}. On the physical car, the desired velocity and steering angle are then actuated by the VESC motor controller. In the simulation, an acceleration limit and delay are used to mimic the motor accelerations on the physical car.

    \subsection{Reward Function}
    The reward function (R) consists of four components designed to enable the car to learn to progress along the track while avoiding obstacles (\(P_O\)) and minimising excessive steering (\(P_S\)) for smooth racing. If a collision occurs with the walls or competing cars, a penalty of -25 (\(P_C\)) applies, and the episode ends. 

    \begin{equation}
        R = R_P(1 - 0.7P_O - 0.3P_ S) - P_C
    \end{equation} 
    
    The primary component is a reward for progress along the track (\(R_P\)). This is computed by determining the closest point to the ego vehicle on the representative track spline. The Euclidean distance between this point and the previously reached point in the prior step is calculated to award progress along the track (\(R_P\)). This method incentivises the agent to drive along the centre of the track. Future work will seek to incorporate a reward that enables racing line-like driving.

    This progression reward is modulated according to obstacle proximity (\(P_O\)) and excessive steering (\(P_S\)). These modifiers are intended to train the car to avoid the walls and opposing cars while minimising excessive steering changes. The logistic sigmoid function penalises excessive steering (\(P_S\)) and unsafe distances from obstacles (\(P_O\)). The final parameters for each component are provided in Table \ref{tab:rewardparameters}.

    \begin{equation}
        P_{O/S} = 1 - \frac{1}{1 + e^{-k(|\Delta \omega| - x_0)}}
    \end{equation}

    \begin{table}[!htp]
        \centering
        \caption{Steering and Obstacle Reward Parameters.}
        \label{tab:rewardparameters}
        \resizebox{\columnwidth}{!}{
        \begin{tabular}{ccc}
        
            \toprule
            Parameter & Steering ($P_S$) & Obstacle ($P_O$)\\
                     \midrule
                     \midrule
            $\Delta \omega$ & Change in Steering Angle & Distance to nearest obstacle min($d_i$)\\
            \midrule
            $x_0$ & Threshold for excessive steering 0.3 & Critical distance threshold 0.5 \\
            \midrule
            $k$ &  15 & 35 \\
            \bottomrule
        \end{tabular}
        }
    \end{table}
    
    The function ensures a gradual increase in penalty, preventing abrupt changes in reward that could destabilise learning. The smooth gradient of the sigmoid aids in stable learning, allowing the agent to adjust its policy effectively without encountering discontinuities. The function assigns minimal penalty when the agent is far from desired conditions while steeply increasing the penalty when it goes over a threshold, discouraging risky behaviour. 
    
    The parameters for each function were tuned based on observations of training the agent without these factors. Steering changes greater than \SI{0.3}{\radian\per\sec} caused unstable driving and frequent crashes, especially in the real world. Penalising the agent for getting within \SI{0.5}{\meter} of the competitors or wall helped enable it to take evasive actions earlier, reducing crashes.

    No specific reward is given for overtaking competitors' vehicles to avoid observations of reward hacking by the agent \cite{DiLangosco2022} in preliminary development. The initial agents were developed with an additional reward for overtaking a competitor. However, this led to poor track navigation performance when no opponent was directly in front of the ego car. The intent is to learn to progress through the track, not simply overtake a competitor. Therefore, this reward component was removed in further training with the obstacle penalty accounting for the competitor.

    \subsection{Learning Agent}
    The learning agent was trained with the TD3 algorithm \cite{Fujimoto2018}. TD3 is well-suited for learning continuous control tasks and has extensive use within simulated and real robotics. The delayed policy updates and target smoothing help prevent overreacting to sudden penalty spikes, which is useful when dealing with obstacle penalties modelled by a sigmoid function. The tuned hyperparameters used in the training are shown in Table \ref{tab:times}. 

    \begin{table}[!htp]
        \centering
        \caption{TD3 Training Parameters.}
    
        \begin{tabular}{cc}
        
            \toprule
            Parameter & Configuration \\
                     \midrule
                     \midrule
            State space & \(v, \theta, d_i\)\\
            & \(i \in 0:10\)\\
            \midrule
            Action space & \(v \in 0:2\)\\
            & \(\theta \in -0.434:0.434\) \\
            \midrule
            Max training steps &  140000 \\
            \midrule
            Max exploration steps & 1000 \\
            \midrule
            Discount factor, \(\gamma\) & 0.95 \\
            \midrule
            Actor learning rate & \(1 \times 10^{-4}\)\\
            \midrule
            Critic learning rate & \(1 \times 10^{-3}\)\\
            \midrule
            Hidden Layers & 256\\
            \bottomrule
        \end{tabular}
        
        \label{tab:times}
    \end{table}

\begin{figure}[!htp]
        \centering
        \includegraphics[width=1\linewidth]{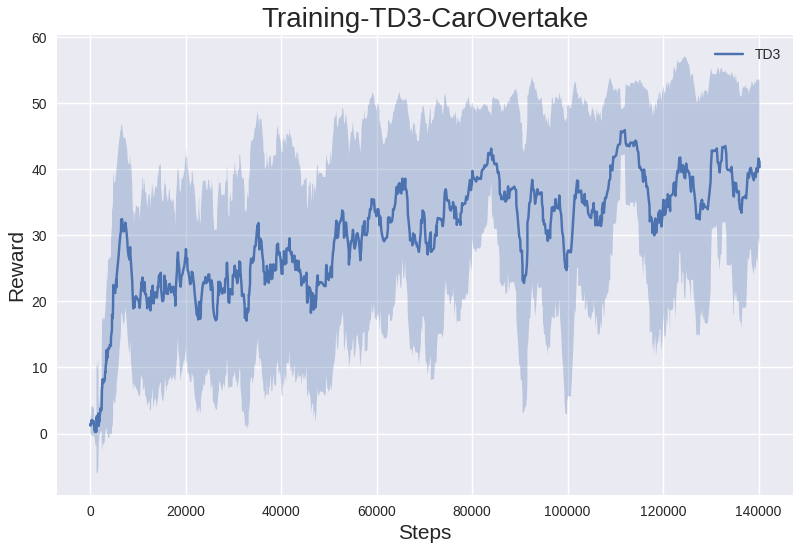}
        \caption{TD3-Overtake training reward over 140,000 training steps in the simulation.}
        \label{fig:training}
    \end{figure}

    The TD3 agent quickly learnt how to navigate the environment and overtake its opponents. The results of training show a steady improvement in racing as shown in Figure \ref{fig:training}. The reward began to plateau after approximately 120,000 steps because the agent would regularly reach the maximum number of steps for the episode and not be able to obtain more reward.

    \subsection{Code}
    All code for reproducing the training and real-world experiments is available here: \url{https://github.com/UoA-CARES/autonomous_f1tenth.git}.

\section{EXPERIMENTS AND RESULTS}

The trained overtaking method, TD3-Overtake, was evaluated in simulation and on a real-world race track against a competitor. The evaluation considers the time required to overtake and whether the cars collide.

\subsection{SIMULATION}
The trained overtaking model (TD3-Overtake) was evaluated in a simulated overtaking scenario. A TD3 model trained without competing cars (TD3-Race) was also evaluated to provide a baseline comparison. Comparing their performance helps determine whether any overtaking behaviour is due to competition awareness or general track adaptation skills.

\begin{figure}[!htp]
    \centering
    \includegraphics[width=1\linewidth]{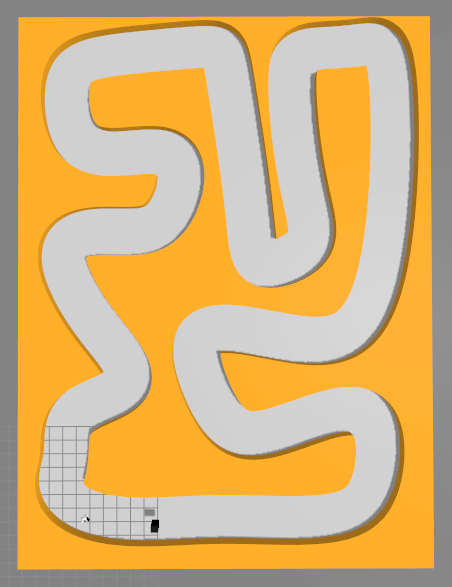}
    \caption{Simulation testing track with track width of \SI{2.5}{\meter}.}
    \label{fig:simtrack}
\end{figure}

The models were tested on a new track (shown in Figure \ref{fig:simtrack}), unknown to either agent, against a single competitor vehicle. Similar to prior work the chasing vehicles had a higher maximum velocity to support overtaking at \SI{2}{\meter\per\sec} vs the competitor vehicle at \SI{1.5}{\meter\per\sec}. 

The two models competed against vehicles running Follow The Gap, TD3-Race, and TD3-Overtake. Each model completed 100 runs against each competitor algorithm (300 per model in total), spawning on the track with the competitor vehicle located eight waypoints ahead (up to \SI{8}{\meter} ahead depending on the curve of the track). The ego vehicle then had 150 steps to make a successful overtake. If an overtake was successful or there was a collision, then the episode would end. 

\begin{figure}[!htp]
    \centering
    \includegraphics[width=1\linewidth]{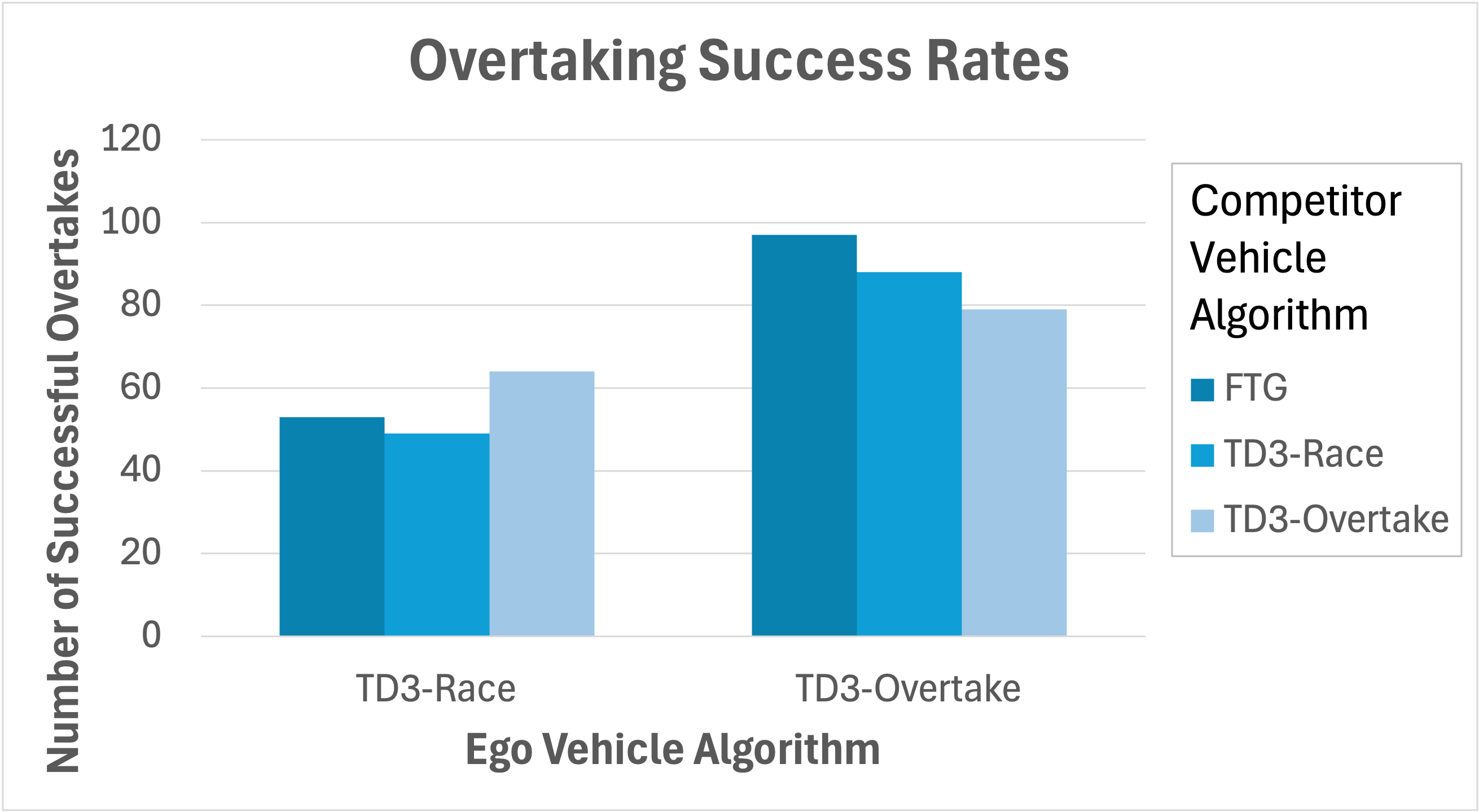}
    \caption{Graph of simulation overtaking results. Each ego vehicle algorithm had 100 attempts to overtake each competitor algorithm.}
    \label{fig:sim_results}
\end{figure}

The overtaking results are shown in Figure \ref{fig:sim_results}. Neither of the models reached the 150-step maximum in any run; they either successfully overtook or crashed before this point. TD3-Overtake had significantly more successful overtakes than TD3-Race, achieving a total success rate of 88\% compared to TD3-Race's success rate of 55\%. These results demonstrate that the training with competing cars has enabled the agent to develop overtaking behaviours. The agent is not simply overtaking due to the higher velocity. 

\subsection{REALITY}
\begin{figure}[!htp]
        \centering
        \includegraphics[width=1\linewidth]{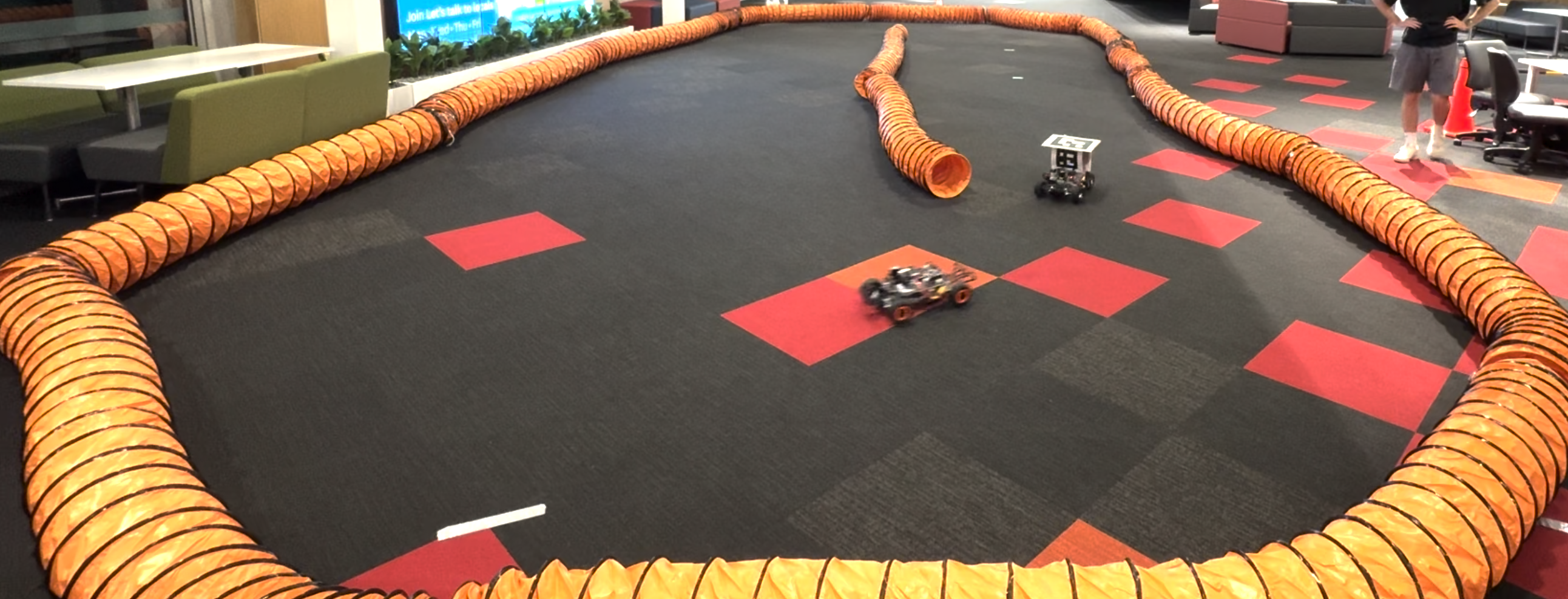}
        \caption{The real-world race track consisting of orange ducting and two competing F1Tenth Race Cars.  The track is made to F1Tenth specifications with maximum dimensions \SI{9}{\meter} x \SI{4.5}{\meter} with a track width of \SI{2}{\meter}.}
        \label{fig:testing}
    \end{figure}

After simulation training, the trained agent was deployed on a real F1Tenth car in a real track, as shown in Figure \ref{fig:testing}. The agent competed against a competitor running either TD3-Overtake, TD3-Race, or Follow the Gap. Due to the end-to-end nature of the reinforcement learning algorithm, no significant changes were required between running the TD3 algorithm on the simulated or physical car. The only change required was a small decrease in maximum speed from \SI{2}{\meter\per\sec} to \SI{1.5}{\meter\per\sec}. This was necessary to account for undesirable network latency causing delays in steering response.
    
\begin{figure}[!htp]
    \centering
    \includegraphics[width=1\linewidth]{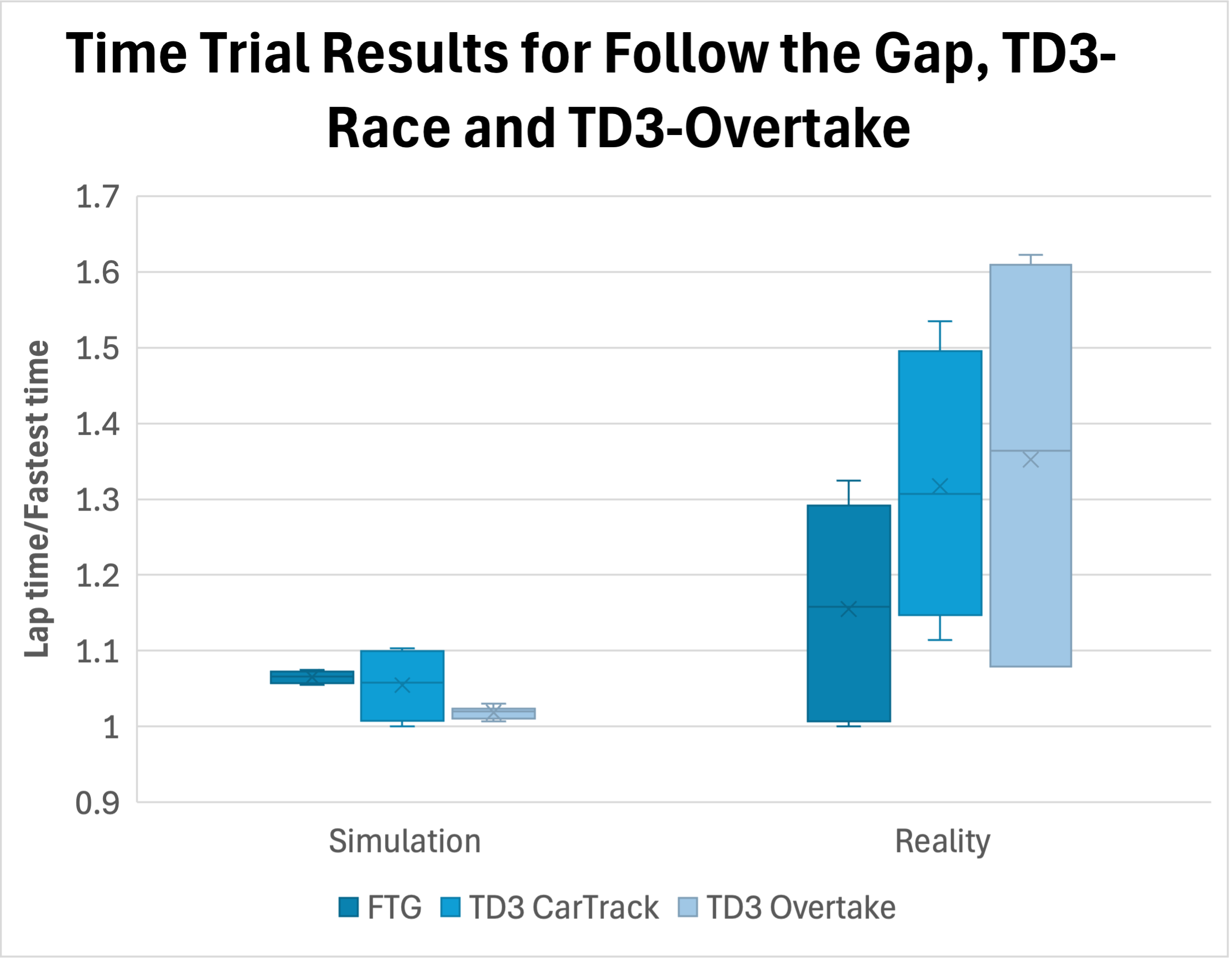}
    \caption{Graph of Time Trial results.}
    \label{fig:irl_results}
\end{figure}

The track used for real-world evaluation was compliant with the F1Tenth competition rules and is shown in Figure \ref{fig:testing}. The agent was tested in a time trial and an overtaking scenario. The time trial component was evaluated to ensure the end-to-end agent had generalised well enough to complete both events at an F1Tenth competition. The time trial consisted of three laps with individual lap times and total times recorded. Crashes would result in a failed attempt. The overtaking agent was compared to TD3-Race and Follow The Gap, as shown in Figure \ref{fig:irl_results}. The results are shown for the real-world testing and a simulation time trial. The results demonstrate that the learned behaviour is capable of producing competitive lap times; however, it is less consistent than TD3-Race, and both TD3 algorithms achieve slower average times than Follow The Gap. Given the relatively similar time performance, without a speed boost, the opportunity to overtake would not be frequent.

\begin{figure}[!htp]
    \centering
    \includegraphics[width=1\linewidth]{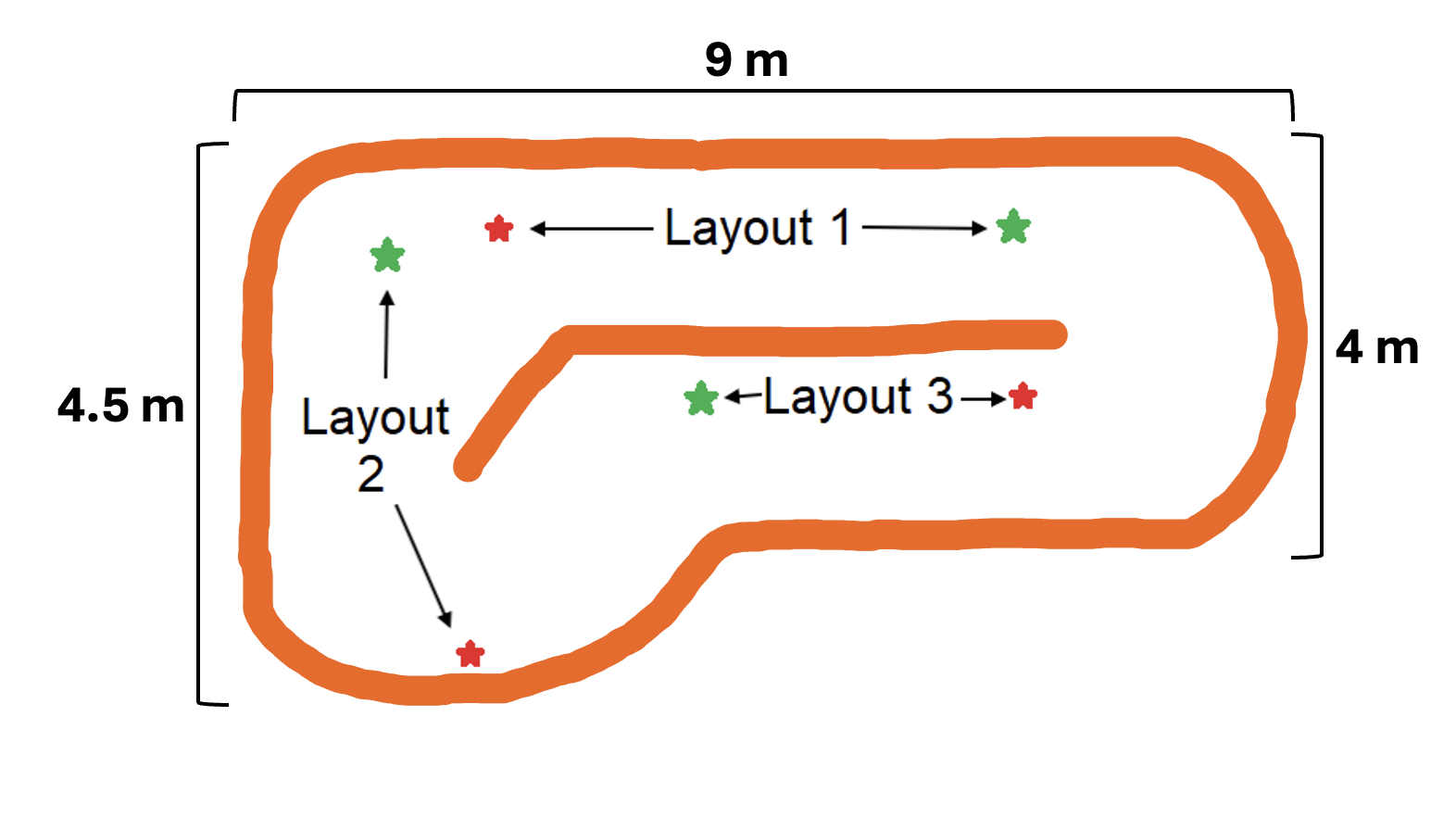}
    \caption{Track evaluation set up where red star is the ego car and green star is the competitor car.}
    \label{fig:enter-label}
\end{figure}

The overtaking scenario involved the ego vehicle starting the lap \SIrange{2}{3}{\meter} from a competitor vehicle. Figure \ref{fig:enter-label} shows the starting locations and track configuration. Video of the overtaking scenario can be found at \url{https://youtu.be/6vRRWeZTG-k}.

\begin{figure}[!htp]
    \centering
    \includegraphics[width=1\linewidth]{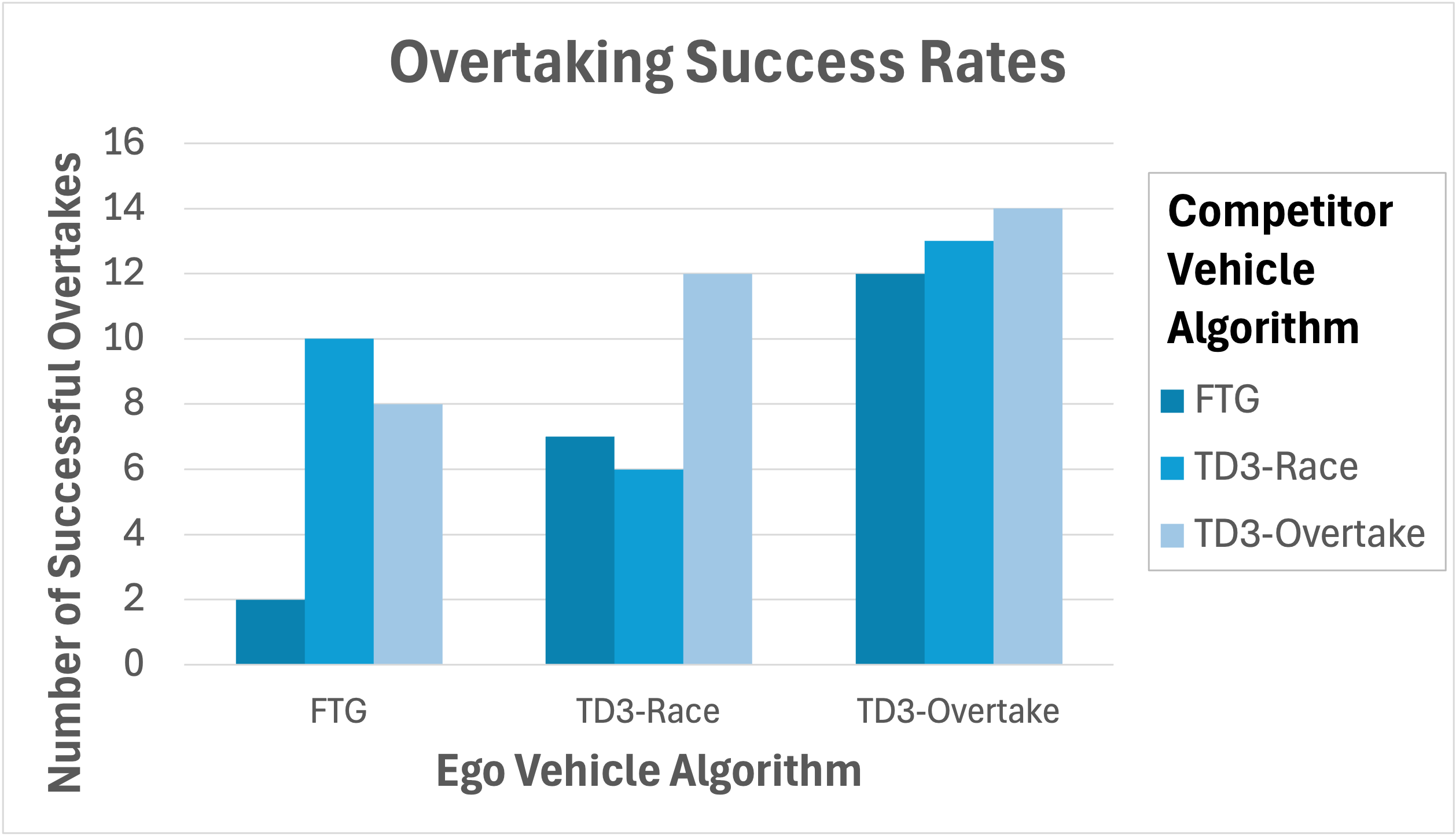}
    \caption{Graph of Overtaking results in reality.}
    \label{fig:overtake_results}
\end{figure}

The competitor vehicle was slowed to a maximum speed of \SI{0.75}{\meter\per\sec} to enable interaction with the ego vehicle. The ego vehicle then had one lap to overtake the competitor vehicle. Crashing or finishing the one lap without a successful overtake would result in a failed attempt. Successful overtakes, unsuccessful overtakes (i.e. attempts where the car failed to pass but did not crash), and the time to overtake were recorded. TD3-Overtake, TD3-Race, and Follow The Gap were tested against each other. The results, shown in Figure \ref{fig:overtake_results}, show that TD3-Overtake performed consistently against all three competitor algorithms, while Follow The Gap and and TD3-Race had varying success against the different algorithms. TD3-Overtake achieved an overall success rate of 87\%, while TD3-Race and Follow The Gap achieved success rates of 56\% and 44\%, respectively.

\section{DISCUSSION}

A significant challenge defined in the literature for developing autonomous overtaking using reinforcement learning is overcoming the Sim-to-Real gap. The results gathered in this study clearly show that the Sim-to-Real gap has been largely overcome in this research through the use of a high-fidelity model and careful parameter selection. This is evidenced by the successful overtaking completed in reality by TD3-Overtake without any real-world training. However, there was a minor decrease in performance in terms of maximum speed, which had to be decreased by 0.5 m/s for the real-world scenarios. This discrepancy may be further minimised by modelling the real car's steering latency within the simulator.

\begin{figure}[!htp]
    \centering
    \includegraphics[width=1\linewidth]{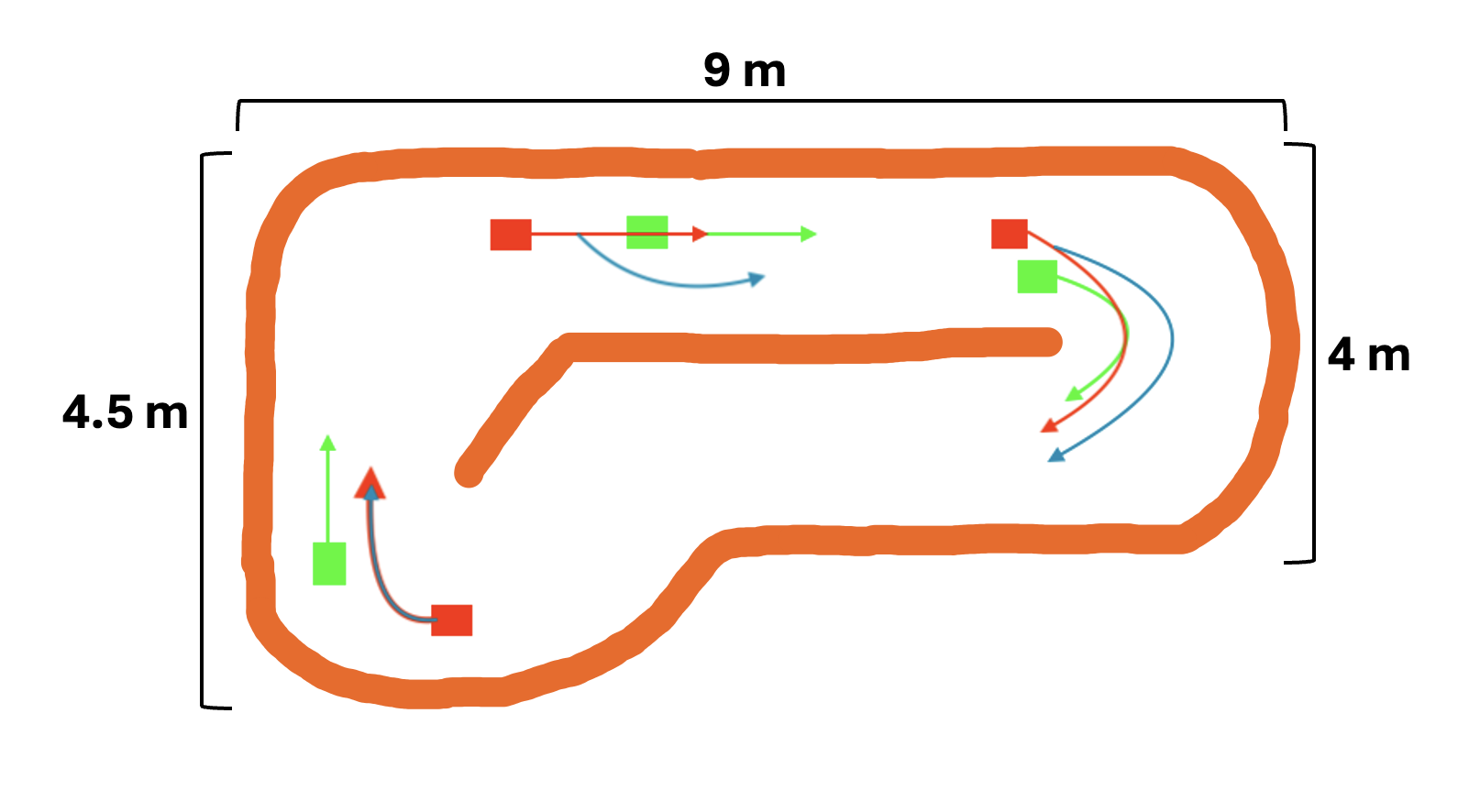}
    \caption{Different trajectories of driving algorithms. The green arrow represents the competitor car's trajectory. The red arrow shows the ego car's trajectory using TD3-Race. Where the red and green arrows intersect, the ego car would either slow down, allowing the competitor car to stay ahead or cause a crash. The blue arrow shows the trajectory taken by TD3-Overtake. }
    \label{fig:overtaking}
\end{figure}

TD3-Overtake learned superior collision avoidance and overtaking skills compared to TD3-Race. The behaviour observed resembled overtaking mechanisms similar to human driving, with careful driving around corners and opportunistic overtaking. The TD3-Overtake model was also the only model that performed well when racing another car running the same model. This is because it can deviate from its regular driving lines to execute overtaking. The TD3-Race and Follow The Gap algorithms mainly achieved overtaking in situations where the competitor car happened to follow a different racing line from the ego car, creating a simple opportunity for the ego car to overtake without deviating from its intended path. This is illustrated in Figure \ref{fig:overtaking}.

In the real-world results, it is noticeable that all methods performed better against the reinforcement learning trained opponent than the Follow The Gap opponent. Qualitatively, this was observed because as the ego vehicle passed the competitor vehicle, the competitor vehicle (running TD3-Race or TD3-Overtake) contributed to avoiding collisions. 

One point of interest from the overtaking scenario is that the TD3-Overtake model generalised its overtaking behaviour for different competitor algorithms. The model was trained to compete only against Follow The Gap. However, it successfully overtook both TD3-Race and TD3-Overtake in addition to Follow The Gap. This shows that it did not just learn how to exploit Follow The Gap's driving style but instead learnt a general ability to overtake a vehicle. This allowed the ego vehicle to overtake its competitor at different points of the track and in different ways. This makes it potentially more robust than traditional overtaking methods, which are limited in overtaking manoeuvres or rely on specific situations, such as straights or certain competitor vehicle positions, to make overtakes.

\section{CONCLUSIONS AND FUTURE WORK}

The study showed that the overtaking model managed to learn to overtake and compete in the simulation and in the real world. The overtaking model performed overtaking manoeuvres against various opponents significantly more consistently than the TD3-Race and Follow the Gap algorithms. End-to-end reinforcement learning minimised discrepancies between the simulation and reality, allowing the model to overcome the sim-to-real gap with minimal changes.

The comparison between TD3-Overtake and TD3-Race clearly shows that the inclusion of opponents in the training environment helped TD3-Overtake to perform better at overtaking manoeuvres than an agent trained without opponents. This is particularly apparent when the algorithms attempt to overtake themselves. TD3-Overtake is able to adapt its strategy to complete the overtake whereas TD3-Race followed the same racing line as the competitor car, leading to slowing down and collisions.

A limitation of the work is the maximum speed of the model. While the model's overtaking abilities outperform other methods in the literature at the same relative speed, the speed at which the model can race is not yet fast enough to compete. This could be remedied by training the model at higher maximum speeds or employing hybrid methods to combine the overtaking model with a higher-speed racing algorithm. 

Some areas have been identified as future work to further build upon the work completed in this research. The first is to increase the maximum speed at which the reinforcement learning model can successfully compete in simulation and the real world. The second is to test and evaluate the overtaking agent's abilities against a broader range of competitor algorithms, such as MPC and Pure Pursuit, and algorithms that actively attempt to block the ego car. The final area of future work is to investigate hybrid methods to combine a strong time-trialling model with a strong overtaking model for maximum racing performance.

\section*{ACKNOWLEDGMENT}
    This research was supported by New Zealand's Science For Technical Innovation (SfTI) on contract UoA3727019.
    
\bibliography{Bibliography}

\begin{thebibliography}{10}
\providecommand{\url}[1]{#1}
\csname url@rmstyle\endcsname
\providecommand{\newblock}{\relax}
\providecommand{\bibinfo}[2]{#2}
\providecommand\BIBentrySTDinterwordspacing{\spaceskip=0pt\relax}
\providecommand\BIBentryALTinterwordstretchfactor{4}
\providecommand\BIBentryALTinterwordspacing{\spaceskip=\fontdimen2\font plus
\BIBentryALTinterwordstretchfactor\fontdimen3\font minus \fontdimen4\font\relax}
\providecommand\BIBforeignlanguage[2]{{%
\expandafter\ifx\csname l@#1\endcsname\relax
\typeout{** WARNING: IEEEtran.bst: No hyphenation pattern has been}%
\typeout{** loaded for the language `#1'. Using the pattern for}%
\typeout{** the default language instead.}%
\else
\language=\csname l@#1\endcsname
\fi
#2}}

\bibitem{Wurman2022}
P.~R. Wurman, S.~Barrett, K.~Kawamoto, J.~MacGlashan, K.~Subramanian, T.~J. Walsh, R.~Capobianco, A.~Devlic, F.~Eckert, F.~Fuchs, \emph{et~al.}, ``Outracing champion gran turismo drivers with deep reinforcement learning,'' \emph{Nature}, vol. 602, no. 7896, pp. 223--228, 2022.

\bibitem{Betz2019}
J.~Betz, A.~Wischnewski, A.~Heilmeier, F.~Nobis, T.~Stahl, L.~Hermansdorfer, B.~Lohmann, and M.~Lienkamp, ``What can we learn from autonomous level-5 motorsport?'' in \emph{9th International Munich Chassis Symposium 2018: chassis. tech plus}.\hskip 1em plus 0.5em minus 0.4em\relax Springer, 2019, pp. 123--146.

\bibitem{OKelly2020}
M.~O'Kelly, H.~Zheng, D.~Karthik, and R.~Mangharam, ``F1tenth: An open-source evaluation environment for continuous control and reinforcement learning,'' \emph{Proceedings of Machine Learning Research}, vol. 123, 2020.

\bibitem{Betz2022}
J.~Betz, H.~Zheng, A.~Liniger, U.~Rosolia, P.~Karle, M.~Behl, V.~Krovi, and R.~Mangharam, ``Autonomous vehicles on the edge: A survey on autonomous vehicle racing,'' \emph{IEEE Open Journal of Intelligent Transportation Systems}, vol.~3, pp. 458--488, 2022.

\bibitem{Murgovski2015}
N.~Murgovski and J.~Sj{\"o}berg, ``Predictive cruise control with autonomous overtaking,'' in \emph{2015 54th IEEE Conference on Decision and Control (CDC)}.\hskip 1em plus 0.5em minus 0.4em\relax IEEE, 2015, pp. 644--649.

\bibitem{Perez2011}
J.~Perez, V.~Milanes, E.~Onieva, J.~Godoy, and J.~Alonso, ``Longitudinal fuzzy control for autonomous overtaking,'' in \emph{2011 IEEE international conference on mechatronics}.\hskip 1em plus 0.5em minus 0.4em\relax IEEE, 2011, pp. 188--193.

\bibitem{Bruedigam2021}
T.~Br{\"u}digam, A.~Capone, S.~Hirche, D.~Wollherr, and M.~Leibold, ``Gaussian process-based stochastic model predictive control for overtaking in autonomous racing,'' \emph{arXiv preprint arXiv:2105.12236}, 2021.

\bibitem{Schwarting2021}
W.~Schwarting, A.~Pierson, S.~Karaman, and D.~Rus, ``Stochastic dynamic games in belief space,'' \emph{IEEE Transactions on Robotics}, vol.~37, no.~6, pp. 2157--2172, 2021.

\bibitem{Kalaria2021}
D.~Kalaria, P.~Maheshwari, A.~Jha, A.~K. Issar, D.~Chakravarty, S.~Anwar, and A.~Towar, ``Local nmpc on global optimised path for autonomous racing,'' \emph{arXiv preprint arXiv:2109.07105}, 2021.

\bibitem{Baumann2024}
N.~Baumann, E.~Ghignone, J.~K{\"u}hne, N.~Bastuck, J.~Becker, N.~Imholz, T.~Kr{\"a}nzlin, T.~Y. Lim, M.~L{\"o}tscher, L.~Schwarzenbach, \emph{et~al.}, ``Forzaeth race stack—scaled autonomous head-to-head racing on fully commercial off-the-shelf hardware,'' \emph{Journal of Field Robotics}, 2024.

\bibitem{Zarrar2024}
M.~M. Zarrar, Q.~Weng, B.~Yerjan, A.~Soyyigit, and H.~Yun, ``Tinylidarnet: 2d lidar-based end-to-end deep learning model for f1tenth autonomous racing,'' in \emph{2024 IEEE/RSJ International Conference on Intelligent Robots and Systems (IROS)}.\hskip 1em plus 0.5em minus 0.4em\relax IEEE, 2024, pp. 2878--2884.

\bibitem{Zhang2023}
J.~Zhang and H.-W. Loidl, ``F1tenth: An over-taking algorithm using machine learning,'' in \emph{2023 28th International Conference on Automation and Computing (ICAC)}.\hskip 1em plus 0.5em minus 0.4em\relax IEEE, 2023, pp. 01--06.

\bibitem{Buyval2017}
A.~Buyval, A.~Gabdulin, R.~Mustafin, and I.~Shimchik, ``Deriving overtaking strategy from nonlinear model predictive control for a race car,'' in \emph{2017 IEEE/RSJ international conference on intelligent robots and systems (IROS)}.\hskip 1em plus 0.5em minus 0.4em\relax IEEE, 2017, pp. 2623--2628.

\bibitem{Song2021}
Y.~Song, H.~Lin, E.~Kaufmann, P.~D{\"u}rr, and D.~Scaramuzza, ``Autonomous overtaking in gran turismo sport using curriculum reinforcement learning,'' in \emph{2021 IEEE international conference on robotics and automation (ICRA)}.\hskip 1em plus 0.5em minus 0.4em\relax IEEE, 2021, pp. 9403--9409.

\bibitem{Loiacono2010}
D.~Loiacono, A.~Prete, P.~L. Lanzi, and L.~Cardamone, ``Learning to overtake in torcs using simple reinforcement learning,'' in \emph{IEEE Congress on Evolutionary Computation}.\hskip 1em plus 0.5em minus 0.4em\relax IEEE, 2010, pp. 1--8.

\bibitem{Huang2015}
H.-H. Huang and T.~Wang, ``Learning overtaking and blocking skills in simulated car racing,'' in \emph{2015 IEEE Conference on Computational Intelligence and Games (CIG)}.\hskip 1em plus 0.5em minus 0.4em\relax IEEE, 2015, pp. 439--445.

\bibitem{Dixit2018}
S.~Dixit, S.~Fallah, U.~Montanaro, M.~Dianati, A.~Stevens, F.~Mccullough, and A.~Mouzakitis, ``Trajectory planning and tracking for autonomous overtaking: State-of-the-art and future prospects,'' \emph{Annual Reviews in Control}, vol.~45, pp. 76--86, 2018.

\bibitem{SureshBabu2022}
V.~Suresh~Babu and M.~Behl, ``Threading the needle—overtaking framework for multi-agent autonomous racing,'' \emph{SAE International Journal of Connected and Automated Vehicles}, vol.~5, no.~1, 2022.

\bibitem{Wang2019a}
M.~Wang, Z.~Wang, J.~Talbot, J.~C. Gerdes, and M.~Schwager, ``Game theoretic planning for self-driving cars in competitive scenarios.'' in \emph{Robotics: Science and Systems}, 2019, pp. 1--9.

\bibitem{Steiner2024}
E.~Steiner, A.~Pama, F.~Zhou, K.~Lee, M.~Liarokapis, and H.~Williams, ``Development of a 3d, high-fidelity simulator for autonomous racingof f1tenth cars,'' in \emph{2024 Australasian Conference on Robotics and Automation (ACRA)}, 2024.

\bibitem{Sezer2012}
V.~Sezer and M.~Gokasan, ``A novel obstacle avoidance algorithm:“follow the gap method”,'' \emph{Robotics and Autonomous Systems}, vol.~60, no.~9, pp. 1123--1134, 2012.

\bibitem{Klapalek2021}
J.~Klap{\'a}lek, A.~Nov{\'a}k, M.~Sojka, and Z.~Hanz{\'a}lek, ``Car racing line optimization with genetic algorithm using approximate homeomorphism,'' in \emph{2021 IEEE/RSJ International Conference on Intelligent Robots and Systems (IROS)}.\hskip 1em plus 0.5em minus 0.4em\relax IEEE, 2021, pp. 601--607.

\bibitem{Evans2023}
B.~D. Evans, H.~W. Jordaan, and H.~A. Engelbrecht, ``Comparing deep reinforcement learning architectures for autonomous racing,'' \emph{Machine Learning with Applications}, vol.~14, p. 100496, 2023.

\bibitem{DiLangosco2022}
L.~L. Di~Langosco, J.~Koch, L.~D. Sharkey, J.~Pfau, and D.~Krueger, ``Goal misgeneralization in deep reinforcement learning,'' in \emph{International Conference on Machine Learning}.\hskip 1em plus 0.5em minus 0.4em\relax PMLR, 2022, pp. 12\,004--12\,019.

\bibitem{Fujimoto2018}
S.~Fujimoto, H.~Hoof, and D.~Meger, ``Addressing function approximation error in actor-critic methods,'' in \emph{International conference on machine learning}.\hskip 1em plus 0.5em minus 0.4em\relax PMLR, 2018, pp. 1587--1596.

\end{thebibliography}
\bibliographystyle{IEEEtran.bst}

\end{document}